\pdfoutput=1

\documentclass{article}

\usepackage{microtype}
\usepackage{graphicx}
\usepackage{booktabs}
\usepackage{float}
\usepackage{hyperref}
\usepackage{xurl}

\usepackage[preprint]{icml2026_template/icml2026}
\usepackage{amsmath}
\usepackage{amssymb}
\usepackage[capitalize,noabbrev]{cleveref}

\graphicspath{{figures/}}

\icmltitlerunning{ForecastBench-Sim}

\begin{document}

\twocolumn[
  \icmltitle{ForecastBench-Sim: A Simulated-World Forecasting Benchmark}
  \icmltitlerunning{ForecastBench-Sim}

  \begin{icmlauthorlist}
    \icmlauthor{Jaeho Lee}{fri}
    \icmlauthor{Nick Merrill}{fri,ucb}
    \icmlauthor{Ezra Karger}{fri}
  \end{icmlauthorlist}

  \icmlaffiliation{fri}{Forecasting Research Institute}
  \icmlaffiliation{ucb}{UC Berkeley}

  \icmlcorrespondingauthor{Jaeho Lee}{jaeho\_lee@brown.edu}
  
  \icmlkeywords{forecasting, benchmark, simulated worlds, language models, probabilistic reasoning}

  \vskip 0.3in
]

\printAffiliationsAndNotice{}

\begin{abstract}
Forecasting benchmarks for general-purpose AI systems usually inherit the constraints of the real world: outcomes resolve slowly, tail events are rare, and counterfactual questions are difficult to score. We introduce ForecastBench-Sim, a simulated-world forecasting benchmark built on game rollouts from Freeciv, a turn-based strategy game modelled on the Civilization series. Forecasters receive a fixed world report (a structured snapshot of the current game state) and answer questions about hidden future states; the benchmark then continues the simulation and scores forecasts. Because the world is simulated, the same setup can generate continuous or binary forecasting questions at arbitrary time horizons, paired intervention worlds for conditional or causal questions, and resolved examples of rare or disruptive outcomes. We describe the benchmark pipeline, question families, scoring protocol, and release artifacts, and report validation slices from model evaluations and an anonymized human pilot. ForecastBench-Sim is intended to complement real-world forecasting benchmarks by providing controlled, immediately resolvable tasks for studying probabilistic reasoning under dynamic world states.
\end{abstract}

\section{Introduction}

Forecasting is a natural testbed for evaluating general-purpose AI systems: good forecasts require extracting relevant evidence, reasoning under uncertainty, and assigning calibrated probabilities to future events. Existing forecasting benchmarks such as ForecastBench are valuable, but real-world evaluation has structural limits \citep{karger2024forecastbench}. Outcomes may take weeks or months to resolve, rare events are sparsely observed, and counterfactual questions usually cannot be scored because only one world is realized.

Simulation offers a complementary route. A simulated world can preserve the basic forecasting structure---partial information, path dependence, interacting agents, and hidden future states---while giving the evaluator control over resolution and interventions. ForecastBench-Sim uses Freeciv rollouts as this substrate \citep{freeciv2026}, building on the CivRealm infrastructure for Freeciv-based AI environments \citep{qi2024civrealm}. Each task is defined by a world report (a structured presentation of game state and history shown to forecasters), a set of forecasting questions, and a hidden future produced by continuing the simulation.

ForecastBench-Sim supports binary event forecasts and continuous distributional forecasts over matched world reports, includes comprehension checks for report comprehension, and can produce paired intervention worlds by mutating savegame state before rollout. We report compact validation results from existing model runs and a small anonymized human pilot.

\paragraph{Related work.}
Real-world LLM forecasting benchmarks rely on naturally occurring questions with delayed, single-realization resolution \citep{karger2024forecastbench,cik2024context,requeima2024llmprocesses,halawi2024forecasting}, which introduces well-documented methodological pitfalls \citep{paleka2025pitfalls}. Pastcasting attempts to address this by reusing past events the model has not seen \citep{gao2025pastcasting}; ForecastBench-Sim instead uses a simulation-backed substrate, closer in spirit to controllable multi-agent environments such as CivRealm \citep{qi2024civrealm}. Unlike causal-reasoning benchmarks that score formal inference over toy graphs \citep{jin2023cladder}, ForecastBench-Sim enables matched fork-and-resolve interventional questions over rich evolving worlds, complementing work on consistency and calibration in language-model forecasting \citep{paleka2024consistency,tian2023calibration}.

\section{Benchmark Design}

\begin{figure*}[t]
  \centering
  \includegraphics[width=.8\textwidth]{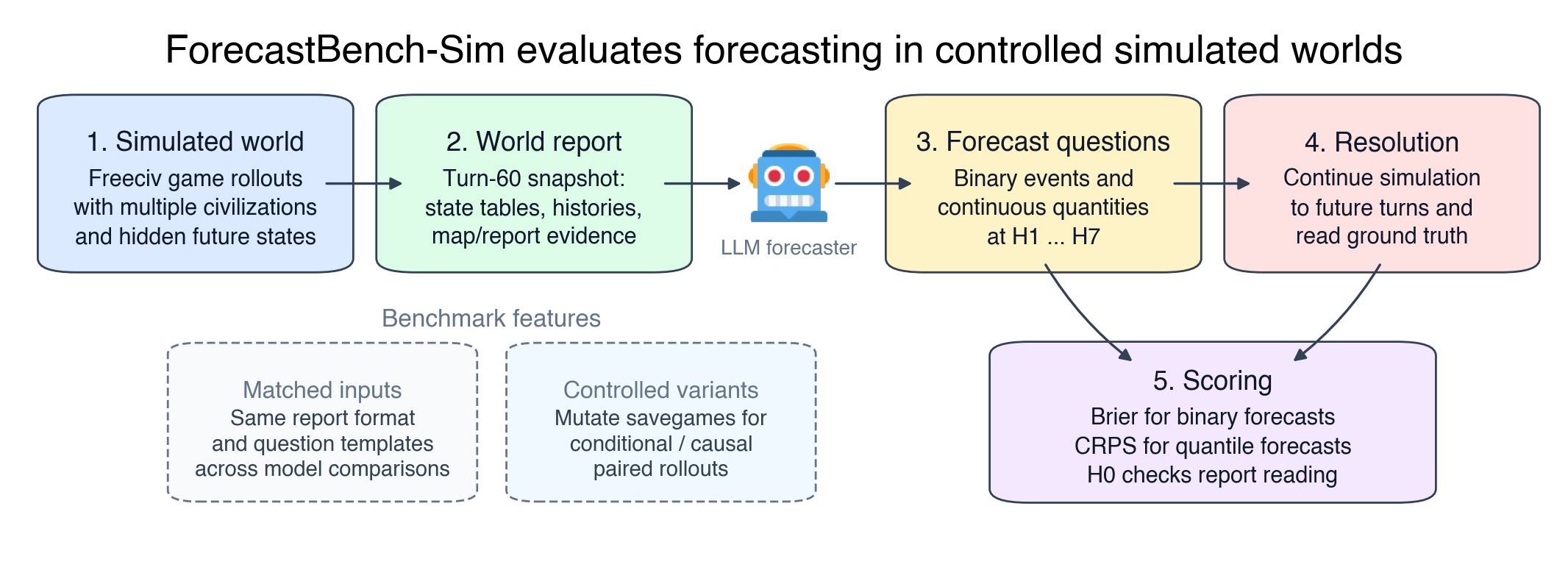}
  \caption{ForecastBench-Sim evaluates forecasts over hidden future states in controlled simulated worlds. A fixed report is shown to forecasters; future turns are withheld until the simulation is continued and scored. The same machinery also supports paired savegame interventions for conditional or causal questions.}
  \label{fig:benchmark-schematic}
\end{figure*}

\subsection{Worlds and Reports}

ForecastBench-Sim begins from Freeciv game rollouts. Each world contains multiple civilizations with evolving cities, technology, territory, treasuries, diplomacy, and conflicts. We expose a fixed snapshot, currently at turn 60, and withhold later turns from the forecaster. The future is generated before scoring but hidden at prediction time.

The forecaster observes a world report rather than the simulator state directly. Reports are designed to be the common input surface for both language models and human participants: they summarize the turn-60 state in structured text, tables, histories, and map/report evidence. This keeps the task close to ordinary forecasting, where the forecaster sees evidence about a world rather than the answer itself, while avoiding retrieval confounds from external web search.

\begin{table}[t]
\centering
\caption{ForecastBench-Sim composition. H0 rows are report-reading checks; H1--H7 rows are forecasting tasks. T/W = templates/worlds.}
\label{tab:benchmark-composition}
\small
\begin{tabular}{lccc}
\toprule
Family & Horiz. & Qs. & T/W \\
\midrule
Binary & H1--H7 & 4,655 & 10 / 11 \\
Continuous & H1--H7 & 2,310 & 6 / 11 \\
H0 binary & H0 & 562 & 10 / 11 \\
H0 continuous & H0 & 165 & 3 / 11 \\
\bottomrule
\end{tabular}
\end{table}

\subsection{Question Families}

From each report, the benchmark generates matched forecasting questions across several future horizons. Binary questions ask whether an event or relation will hold at a later turn, and continuous questions ask for a future value of a world variable. For example, a binary question may ask whether a civilization's treasury will exceed a threshold at a future turn, while the matched continuous question asks for the treasury value itself.

Continuous questions elicit five quantiles, currently p10, p25, p50, p75, and p90. The main continuous families include city counts, technologies, and treasury values; binary templates cover a broader set of comparisons and threshold events. Forward-looking tasks use horizons H1--H7, corresponding to 30-turn increments past the turn-60 snapshot (turns 90 through 270, or roughly the second half of a typical game). H0 questions ask about the snapshot state itself and serve as comprehension checks rather than forecasting tasks.

\subsection{Scoring}

Binary forecasts are scored with Brier score \citep{brier1950verification}; continuous forecasts are scored with CRPS computed from the elicited quantiles \citep{gneiting2007proper}. We normalize CRPS by a fixed per-family value range (cities: 40, technologies: 60, treasury: 2000) so that the three continuous templates can be averaged without changing the per-question scoring rule. These ranges match the bin edges used in the human pilot (\cref{app:human-pilot}), keeping model and human scores on the same scale. H0 checks are reported separately because they test report reading rather than forecasting.

\section{Controlled Evaluation Affordances}

The benchmark supports three forecasting regimes, distinguished by how the conditioning event is established. Unconditional forecasts target $P(Y)$ from a fixed report; observational conditional forecasts target $P(Y \mid X)$ from natural variation across a corpus of simulated worlds; interventional forecasts target $P(Y \mid \mathrm{do}(X))$, where $X$ is set by mutating the savegame before rollout. Each regime uses the same proper scoring rules. The third is the regime that real-world forecasting benchmarks cannot ordinarily resolve, because only one history is realized; the simulator turns it into a scored question by producing a separate ground truth in each branch.

The unconditional benchmark is the closest analogue to existing real-world forecasting benchmarks. A forecaster receives a report and predicts what will later happen, without being told that any special intervention occurred. This setting supports both binary questions, scored by Brier score, and continuous questions, scored by CRPS, yielding a ForecastBench-style evaluation surface with automatic resolution and fixed prompts.

The interventional regime is the one real-world benchmarks cannot ordinarily score: the benchmark copies a savegame at the forecast turn, applies an intervention, rolls out the modified world, and scores forecasts against the intervention-world outcome. Existing pilots use two interventions, switching a civilization's government to Republic and adding 500 gold to its treasury, and reuse the same target templates under baseline and conditional framings. Across the archived nine-model slice, mean intervention gain is positive for every curated model on both pilots (+0.029 to +0.057 on Republic; +0.013 to +0.062 on +500 gold; \cref{fig:intervention-gap}). An Opus 4.5 placebo control leaves Brier nearly unchanged under null-conditional wording (0.165 vs.\ 0.169 baseline) while the real Republic conditional rises to 0.360 (\cref{fig:null-conditional}), suggesting that the response is not explained by conditional phrasing alone.

The observational regime can instead be populated by sampling a corpus of simulated worlds and measuring how often events co-occur. This slice is not yet part of the released artifact, but it remains cleanly separable from the interventional regime because the data source is a corpus rather than a forked rollout.

Finally, simulated worlds make tail-risk evaluation practical because large shocks can be sampled densely instead of waited for. As one example, a companion anonymous analysis on the same question family finds the share of treasury outcomes falling below the model's stated p10 rising to roughly 50\% by H6--H7 (\cref{fig:tail-risk}; \citealp{anonymous2026capability}).

\section{Validation Results}

\begin{figure}[t]
  \centering
  \includegraphics[width=0.95\linewidth]{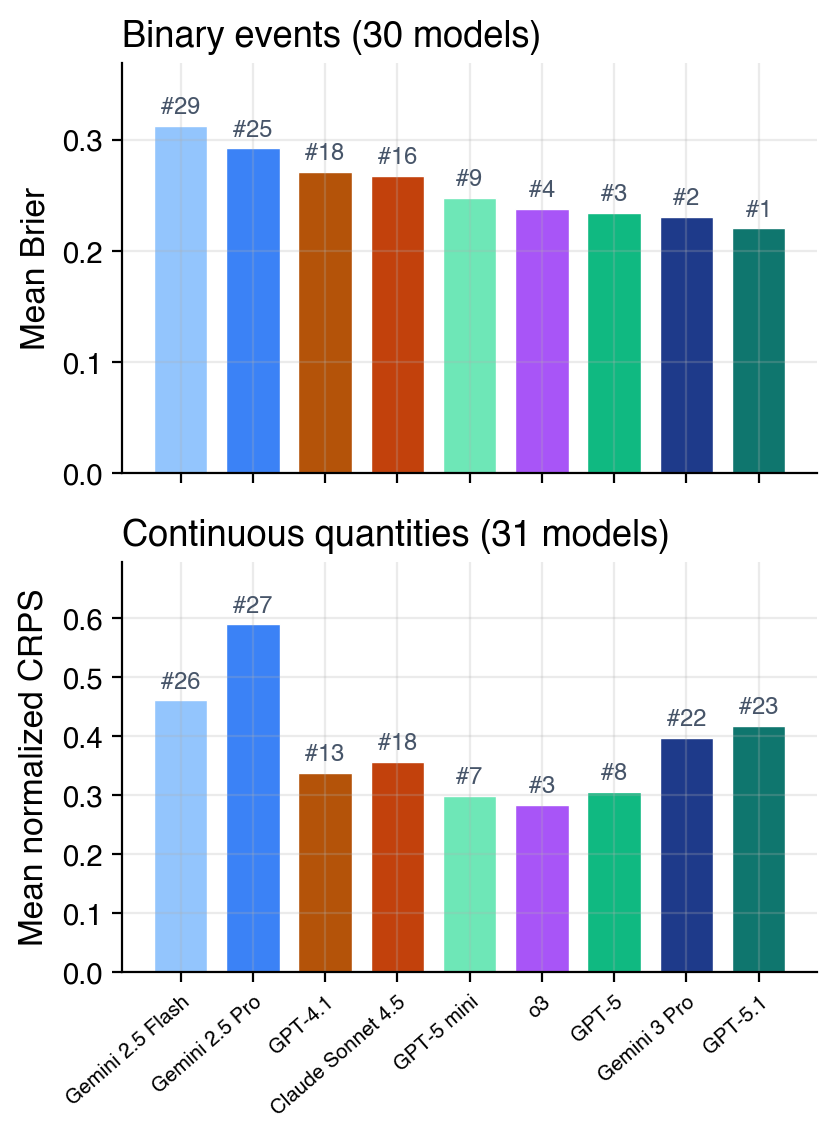}
  \caption{Compact validation slice for nine curated models with full coverage across model, H0, human-baseline, and intervention artifacts. Bars show average forward-looking performance; rank markers give each model's position in the larger binary or continuous run. Binary and continuous scores produce different ordinal rankings of model performance; this is the subject of contemporary work \cite{anonymous2026capability}.}
  \label{fig:model-validation}
\end{figure}

We use existing model runs to check that the benchmark behaves like a forecasting evaluation rather than a report-parsing exercise. The main unconditional run contains 4,655 binary questions across 30 models and 2,310 continuous questions across 31 models. The compact leaderboard in \cref{fig:model-validation} shows the curated subset with complete coverage across the artifacts used in this paper, while preserving rank context from the larger runs. Across the nine curated models, mean Brier on H1--H7 binary questions ranges from 0.220 (GPT-5.1) to 0.313 (Gemini 2.5 Flash), and mean normalized CRPS on H1--H7 continuous questions ranges from 0.283 (o3) to 0.590 (Gemini 2.5 Pro), a spread of roughly 40\% on Brier and a factor of two on CRPS. Averaged across H1--H7, ForecastBench-Sim binary Brier correlates with ForecastBench Dataset Brier at Spearman $\rho = +0.43$ ($p = 0.018$, $N = 30$ overlapping models) and with the Epoch Capabilities Index \citep{ho2025rosetta} at $|\rho| = 0.48$ ($p = 0.007$), indicating that ForecastBench-Sim rankings agree with both a published real-world forecasting benchmark and a general capability index. Per-horizon correlations and the underlying model lists are in \cref{app:capability-correlations}.

\begin{figure}[t]
  \centering
  \includegraphics[width=0.95\linewidth]{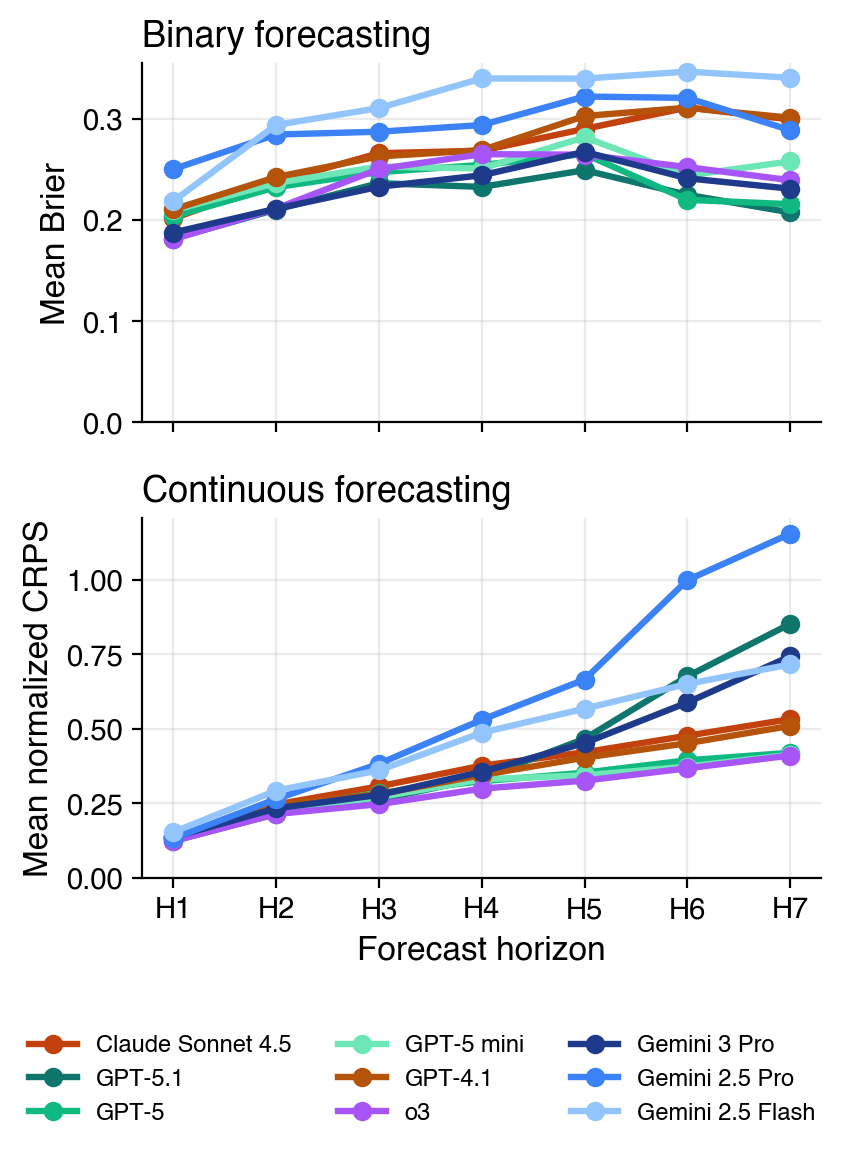}
  \caption{ForecastBench-Sim captures horizon-dependent difficulty: questions about further-future events are more difficult than questions about nearer-future events. (Scores are averaged by horizon for the curated model set.)}
  \label{fig:horizon-curve}
\end{figure}

The benchmark also shows meaningful horizon structure (\cref{fig:horizon-curve}). Averaged over the curated set, mean Brier rises from 0.205 at H1 to 0.264 at H7 (peaking at 0.287 at H5), and mean normalized CRPS rises monotonically from 0.134 at H1 to 0.639 at H7---a 4.8$\times$ increase as the forecast horizon extends from 30 to 210 turns past the snapshot. This indicates that the simulated tasks contain forecastable signal and horizon-dependent uncertainty. We do not intend this workshop paper to be a final model-ranking paper.

H0 checks provide a separate check on the input format. All nine curated models have H0 binary Brier below 0.032 and H0 continuous normalized CRPS below 0.024 (most score 0.000; see \cref{tab:h0-comprehension}), suggesting that forward-looking errors are not primarily caused by inability to read the turn-60 reports. Detailed H0 results are in \cref{app:h0}.

Finally, the anonymized human pilot demonstrates that the same task family can be presented to people. Ten participants answered 24 continuous questions over two worlds, covering city, technology, and treasury templates at H1, H3, H4, and H6. The crowd-mean normalized CRPS is close to a uniform-bin baseline in this small pilot, so we treat it as a simple feasibility check for a larger human baseline rather than as a definitive model comparison; details are in \cref{app:human-pilot}.

\section{Future Work}

ForecastBench-Sim currently uses Freeciv because it provides an inspectable multi-agent world with saved states and replay infrastructure, but the broader goal is a family of simulation-backed slices that plug into the same evaluation interface alongside dataset and market questions in benchmarks like ForecastBench. Freeciv, due to its complexity and wide range of mechanics (as well as forecastable quantities), is likely the best single fit among simulated worlds to assess general forecasting ability. However, it could certainly be supplemented with more specialized simulations—different substrates would expose different forecasting skills. For instance, a question set based on generative-agent social simulators such as Concordia \citep{vezhnevets2023concordia} might probe whether models can predict the behavior of other agents under social pressure. Epidemiological agent-based models would produce dense tail-risk material---outbreak surges and intervention counterfactuals---without waiting for real outbreaks. 

A second extension could change the scoring of forecasts. Instead of resolving a probability against a single $0/1$ outcome, the same starting world can be rolled out $N$ times under different RNG seeds and forecasts resolved against the empirical event frequency across rollouts. This converts resolution from a single noisy realization into a target close to the simulator's true generating probability---only possible because the world is simulated.

Some further directions are natural under the same machinery: sequential forecasting, where the report is revealed at multiple snapshots and models must update as evidence accumulates, and counterfactual coherence checks, where the same target is queried under multiple conditional framings to test whether a model's conditionals respect basic probability axioms \citep{paleka2024consistency}.

\section{Limitations}

ForecastBench-Sim is not a substitute for real-world forecasting benchmarks. Freeciv worlds are simplified and stylized, report formatting can affect both models and humans, and the current human study is only a pilot. The conditional and causal artifacts are demonstrations rather than a fully powered intervention benchmark. The current substrate also uses a fixed ruleset and rule-based AI opponents, so forecasting skill measured here may not transfer cleanly to worlds with adaptive, human-like adversaries.

\section{Conclusion}

ForecastBench-Sim is a simulated-world forecasting benchmark designed to complement real-world evaluations rather than replace them. In the current release, it contributes three main capabilities within a single forecasting interface: unconditional binary and distributional forecasting over fixed world reports, paired intervention rollouts for scored causal-conditional questions, and dense sampling of disruptive outcomes that can support tail-risk analysis with immediate resolution. We view that combination as the benchmark's main contribution: not a final empirical ranking of forecasters, but a reusable evaluation substrate on which later work can make narrower claims about calibration, causal updating, and rare-event forecasting behavior.

\clearpage
\bibliographystyle{icml2026_template/icml2026}
\bibliography{forecastbench_sim}

\clearpage
\appendix
\onecolumn

\clearpage
\section{Sample World Report}
\label{app:sample-report}

Each forecasting task is built around a single \emph{world report}: a structured text document, optionally accompanied by territory map images, that summarizes the state of a Freeciv game at the snapshot turn (currently turn 60). The report is the only world-specific input shown to a forecaster (model or human); the underlying simulator state is never exposed directly. This appendix shows one report (game seed \texttt{seed5}) so readers can see the actual evidence surface that produces the binary and continuous questions.

\begin{figure}[H]
  \centering
  \includegraphics[width=.9\linewidth]{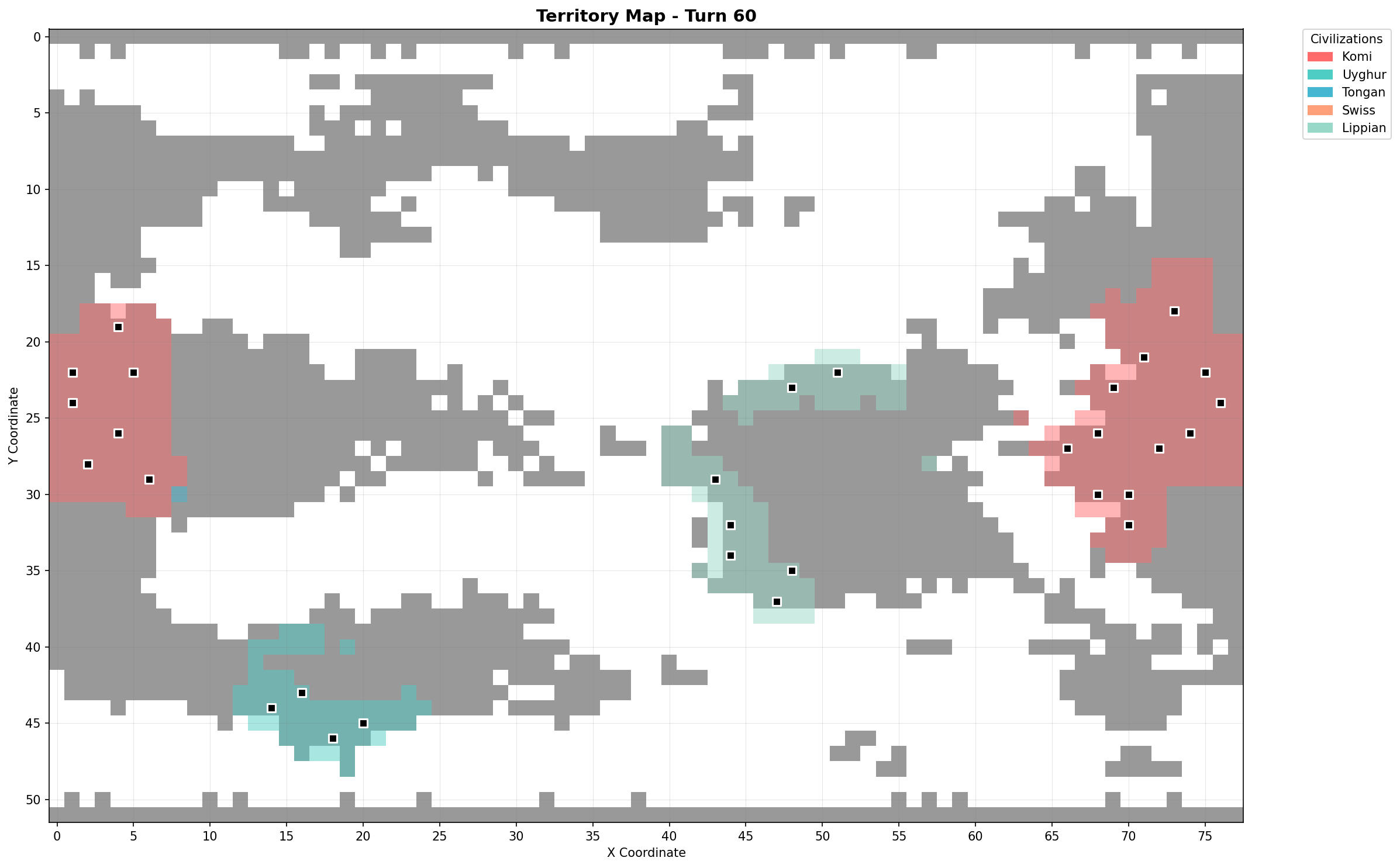}
  \caption{Turn-60 territory map from world \texttt{seed5}. Colored regions are civilization-controlled tiles; black squares are cities; grey is unclaimed land; white is ocean. The full report includes one such map every 10 turns from turn 10 to turn 60.}
  \label{fig:sample-territory-map}
\end{figure}

\paragraph{Territory map.} The report references a sequence of PNG territory maps sampled every ten turns. \Cref{fig:sample-territory-map} shows the turn-60 map for \texttt{seed5}, with colored regions for each civilization's controlled tiles and black squares marking cities. The map makes the gross strategic situation legible at a glance: Komi (red) dominates a coastal block in the northeast with $\sim$19 cities, Lippian (light green) has a smaller but growing central holding, and Uyghur (teal) is confined to a single southern enclave. Two civilizations (Tongan, Swiss) have effectively collapsed and hold no territory.

\paragraph{Structured text.} The text portion of the report is a sequence of labeled tables: a roster of civilizations, the current-turn state vector for each (score, government, treasury, population, techs, cities, territory, military, production), and per-quantity time series sampled every 5 turns from turn 1 to turn 60. The excerpt below shows the opening of the \texttt{seed5} report:

{\footnotesize
\begin{verbatim}
WORLD REPORT TXT v1
Game ID: seed5
Snapshot turn: 60
Map size: 78x52

CIVILIZATIONS
ID | Name       | Nation ID
---+------------+----------
 0 | Komi       | 261
 1 | Uyghur     | 518
 2 | Tongan     | 500
 3 | Swiss      | 479
 4 | Lippian    | 286

CURRENT STATE (TURN 60)
ID | Name    | Score | Gov.      | Treas. | Techs | Cities | Terr.
---+---------+-------+-----------+--------+-------+--------+------
 0 | Komi    | 54    | Despotism | 261    | 10    | 19     | 270
 1 | Uyghur  | -1    | Anarchy   | 0      |  9    |  4     |  55
 2 | Tongan  | -1    | Anarchy   | 0      |  9    |  0     |   1
 3 | Swiss   | 37    | Despotism | 455    |  8    |  0     |   0
 4 | Lippian | 39    | Despotism | 325    |  9    |  7     |  90

TREASURY (sampled every 5 turns; turns 1..60)
0 Komi:    50, 55, 65, 76, 91, 102, 113, 132, 151, 166, 190, 220, 261
1 Uyghur:   0,  0,  0,  0,  0,  98, 114, 131, 149, 179,   0,   0,   0
3 Swiss:    0,  0,  0,  0,  0, 229, 249, 269, 293, 325, 365, 410, 455
4 Lippian:  0,  0,  0,  0,  0,   0,   0, 184, 200, 223, 255, 290, 325

(...analogous blocks for population, techs known, wonders,
 cities count, territory size, science/trade/food/shield
 production, units, military units, scores, rankings...)

EVENTS (chronological; abridged)
 21 | government_change | Uyghur  | Anarchy -> Despotism
 22 | government_change | Swiss   | Anarchy -> Despotism
 32 | government_change | Lippian | Anarchy -> Despotism
 48 | government_change | Uyghur  | Despotism -> Anarchy
 56 | wonder_completed  | Komi    | Great Wall in Syktyvkar
 58 | tech_discovered   | Uyghur  | The Republic
 60 | tech_discovered   | Swiss   | The Republic
 60 | city_founded      | Lippian | Detmold, Lemgo
\end{verbatim}
}

The full \texttt{seed5} report is roughly 300 lines of structured text plus 6 territory map PNGs (turns 10, 20, \ldots, 60). Every continuous and binary question in the benchmark is grounded in a report of this form: for example, the matched H6 continuous and binary questions ``How many cities will Komi have at turn 240?'' and ``Will Komi have more than 25 cities at turn 240?'' are both answerable in principle from the trajectories and event history shown above. A forecaster who can extrapolate Komi's city-founding rate, anticipate the looming republic transitions in Uyghur and Swiss, and reason about the territory map's choke points has the same evidence base whether they are a language model or a human participant.

\clearpage
\section{H0 Report-Comprehension Checks}
\label{app:h0}

H0 questions ask for quantities that are already present in the turn-60 report. They are not included in forward-looking scores, but they are important for interpreting those scores: a model that fails H0 may be failing to extract the report, while a model that passes H0 and fails H1--H7 is more plausibly failing at forecasting. The curated models pass this check on the continuous questions and have low binary Brier scores, so the main validation curves are not driven by obvious report-reading failures.

\begin{table}[H]
\centering
\caption{H0 comprehension checks. H0 questions are answerable from the turn-60 report and test report reading rather than future forecasting. Lower is better.}
\label{tab:h0-comprehension}
\small
\resizebox{0.78\linewidth}{!}{%
\begin{tabular}{lrrrr}
\toprule
Model & Bin. Brier & Bin. $n$ & Cont. nCRPS & Cont. $n$ \\
\midrule
Claude Sonnet 4.5 & 0.0142 & 562 & 0.0000 & 165/165 \\
GPT-5.1 & 0.0178 & 562 & 0.0008 & 165/165 \\
GPT-5 & 0.0035 & 562 & 0.0000 & 165/165 \\
GPT-5 mini & 0.0002 & 562 & 0.0004 & 165/165 \\
GPT-4.1 & 0.0276 & 562 & 0.0001 & 165/165 \\
o3 & 0.0053 & 562 & 0.0029 & 165/165 \\
Gemini 3 Pro & 0.0319 & 562 & 0.0240 & 165/165 \\
Gemini 2.5 Pro & 0.0047 & 562 & 0.0000 & 165/165 \\
Gemini 2.5 Flash & 0.0018 & 562 & 0.0000 & 165/165 \\
\bottomrule
\end{tabular}
}
\end{table}

\clearpage
\section{Human Pilot Comparison}
\label{app:human-pilot}

The anonymized pilot contains 10 participants, 24 continuous questions, and 240 individual forecasts. Participants saw two simulated-world reports and allocated probability mass across five bins for city, technology, and treasury questions at H1, H3, H4, and H6. The released data remove raw survey identifiers, timestamps, demographics, consent fields, free-text comments, and mouse/click events.

The pilot is useful for checking task feasibility and for constructing a preliminary human reference point. Mean individual normalized CRPS is 0.171, the crowd mean is 0.154, and the uniform-bin baseline is 0.152 on these 24 questions. The crowd aggregation helps relative to individual responses, but the pilot does not clearly beat the uninformed baseline. This is consistent with the study being small, general-population, and cognitively demanding; it motivates a larger and more carefully powered human baseline rather than supporting a strong human-vs-model claim.

\begin{figure}[H]
  \centering
  \includegraphics[width=.72\textwidth]{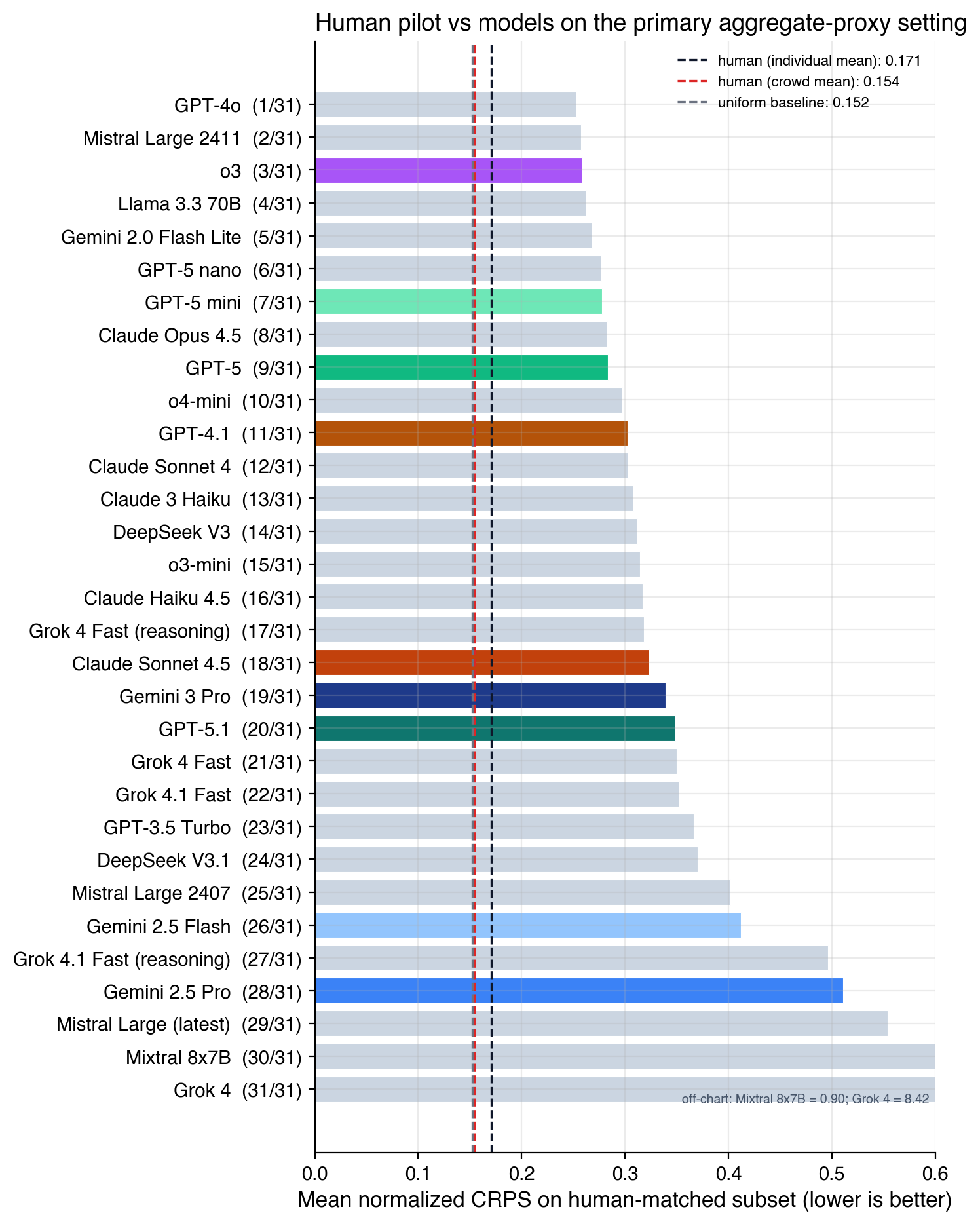}
  \caption{Human pilot compared with models on the closest aggregate proxy: cities, technologies, and treasury at H1, H3, H4, and H6. The human pilot has 24 questions on 2 worlds; the model proxy uses the same template/horizon family across the larger model question set, so this is not a matched head-to-head comparison.}
  \label{fig:human-proxy}
\end{figure}

\clearpage
\section{Capability and Cross-Benchmark Correlations}
\label{app:capability-correlations}

The main text reports two pooled-horizon correlations indicating that ForecastBench-Sim rankings track both general capability (the Epoch Capabilities Index, ECI; \citealp{ho2025rosetta}) and a published real-world forecasting benchmark (ForecastBench Dataset Brier; \citealp{karger2024forecastbench}). This appendix expands those numbers with per-horizon Spearman correlations and the underlying model lists.

\paragraph{Data sources.} ECI scores are taken from the Epoch AI benchmarking hub \citep{epoch2026benchmarks}. ForecastBench Dataset Brier scores are zero-shot entries from \texttt{leaderboard\_baseline.csv}, restricted to dataset questions (autoresolving against published time series) and excluding human-judged market questions; when a model family has multiple zero-shot entries, we use the entry with the largest forecast count. After matching by model identifier, 30 models overlap between ForecastBench-Sim's binary run and the ForecastBench leaderboard, and all 31 binary-run models have ECI scores.

\paragraph{Per-horizon correlations.} \Cref{tab:capability-correlations} reports Spearman $\rho$ between ForecastBench-Sim binary Brier and the two external measures, by horizon. Both correlations are sign-adjusted to ``pro-g'' convention (positive $\rho$ = more capable models do better) for readability: ForecastBench-Sim Brier and FB Dataset Brier are both lower-is-better, so the raw correlation is positive and we leave it as is; ECI is higher-is-better against a lower-is-better Brier, so we negate the raw correlation. Significance markers use $p < 0.05$ (*), $p < 0.01$ (**), $p < 0.001$ (***).

\begin{table}[H]
\centering
\caption{Spearman correlation between ForecastBench-Sim binary Brier and external measures, by forecast horizon. ECI column uses $N{=}31$ models; FB Dataset column uses $N{=}30$ models that overlap with the ForecastBench leaderboard. Both correlations are sign-adjusted so that positive values indicate more capable models perform better.}
\label{tab:capability-correlations}
\small
\begin{tabular}{lcccc}
\toprule
Horizon & $\rho$(Brier, ECI) & $p$ & $\rho$(Brier, FB Dataset) & $p$ \\
\midrule
H0 & $+0.551$ & $0.002$ ** & --- & --- \\
H1 & $+0.634$ & $0.0001$ *** & $+0.630$ & $0.0002$ *** \\
H2 & $+0.433$ & $0.015$ * & $+0.433$ & $0.017$ * \\
H3 & $+0.420$ & $0.019$ * & $+0.334$ & $0.072$ \\
H4 & $+0.338$ & $0.063$ & $+0.241$ & $0.199$ \\
H5 & $+0.294$ & $0.109$ & $+0.172$ & $0.363$ \\
H6 & $+0.456$ & $0.010$ * & $+0.361$ & $0.050$ * \\
H7 & $+0.560$ & $0.001$ ** & $+0.477$ & $0.008$ ** \\
\midrule
Pooled H1--H7 & $+0.479$ & $0.007$ ** & $+0.431$ & $0.018$ * \\
\bottomrule
\end{tabular}
\end{table}

The H1 row is the cleanest single-horizon comparison and shows the strongest signal: ForecastBench-Sim H1 binary Brier correlates with ECI at $|\rho|=0.634$, essentially matching ForecastBench's own internal capability correlation $|\rho|=0.681$ (N{=}23 FB-leaderboard models with ECI scores) and with ForecastBench Dataset Brier itself at $\rho=+0.630$. Pooling across all forward horizons attenuates both correlations modestly but leaves them significant. We do not report continuous-side correlations here; those are the subject of a companion anonymous analysis \citep{anonymous2026capability}.

\paragraph{Model coverage.} The 30-model FB-overlap set spans Anthropic (Claude 3 Haiku; Claude 4 Sonnet, Haiku 4.5, Opus 4.5, Sonnet 4.5), OpenAI (GPT-3.5 Turbo, GPT-4o, GPT-4.1, GPT-5, GPT-5 mini, GPT-5 nano, GPT-5.1, o3, o3-mini, o4-mini), Google (Gemini 2.0 Flash Lite, Gemini 2.5 Flash, Gemini 2.5 Pro, Gemini 3 Pro), Mistral (Large 2407/2411/latest), xAI (Grok-4, Grok-4-fast reasoning/non-reasoning, Grok-4.1-fast reasoning/non-reasoning), and open-weights models served via Together AI (DeepSeek V3, V3.1; Llama 3.3 70B; Mixtral 8x7B). The ECI-only set adds Claude Opus 4.6, which is not yet on the ForecastBench leaderboard.

\clearpage
\section{Full Validation Diagnostics}
\label{app:additional-validation}

The main text uses compact figures to keep the benchmark-design narrative readable. This appendix section expands those results with a full leaderboard, a metric-comparison view, template-level slices, and calibration diagnostics.

\begin{figure}[H]
  \centering
  \includegraphics[width=.72\textwidth]{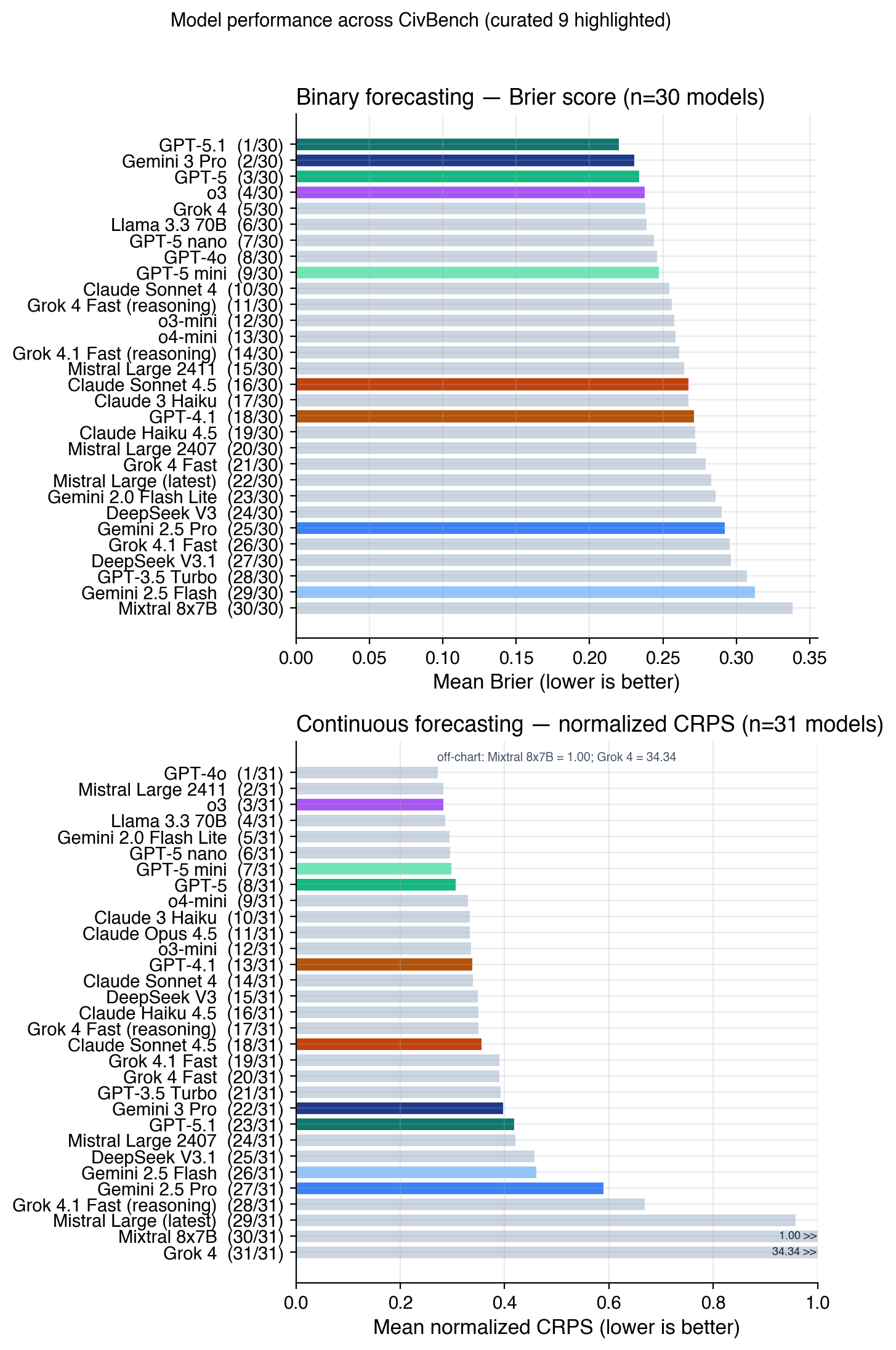}
  \caption{Full binary and continuous leaderboard. Curated models are highlighted, while additional models provide context for the compact validation figure in the main text.}
  \label{fig:full-leaderboard}
\end{figure}

\begin{figure}[H]
  \centering
  \includegraphics[width=.68\textwidth]{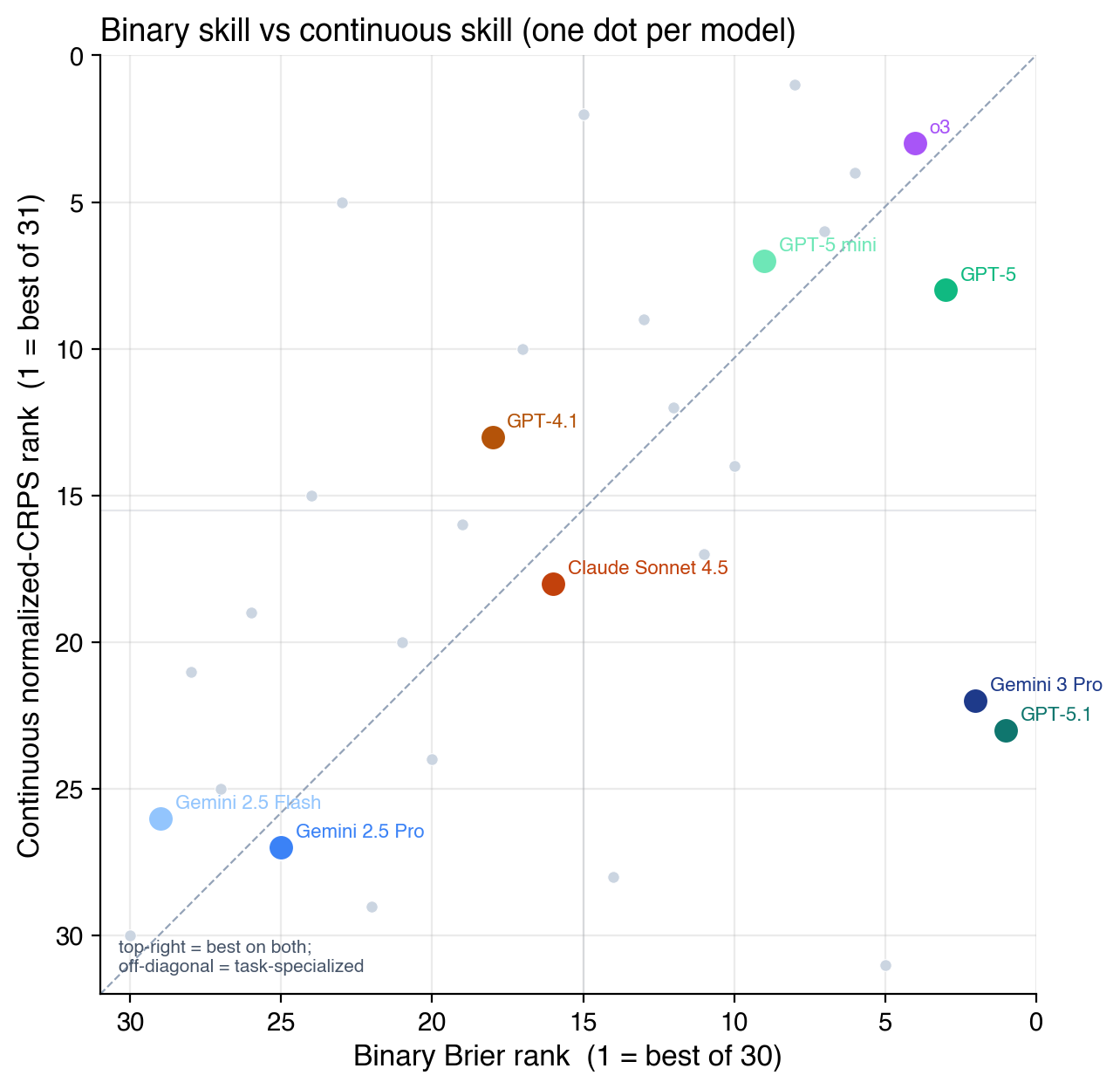}
  \caption{Binary and continuous performance are related but not identical. Some models rank well under Brier score while ranking lower under normalized CRPS, consistent with the benchmark measuring more than one forecasting skill.}
  \label{fig:binary-continuous}
\end{figure}

\begin{figure}[H]
  \centering
  \includegraphics[width=.95\textwidth]{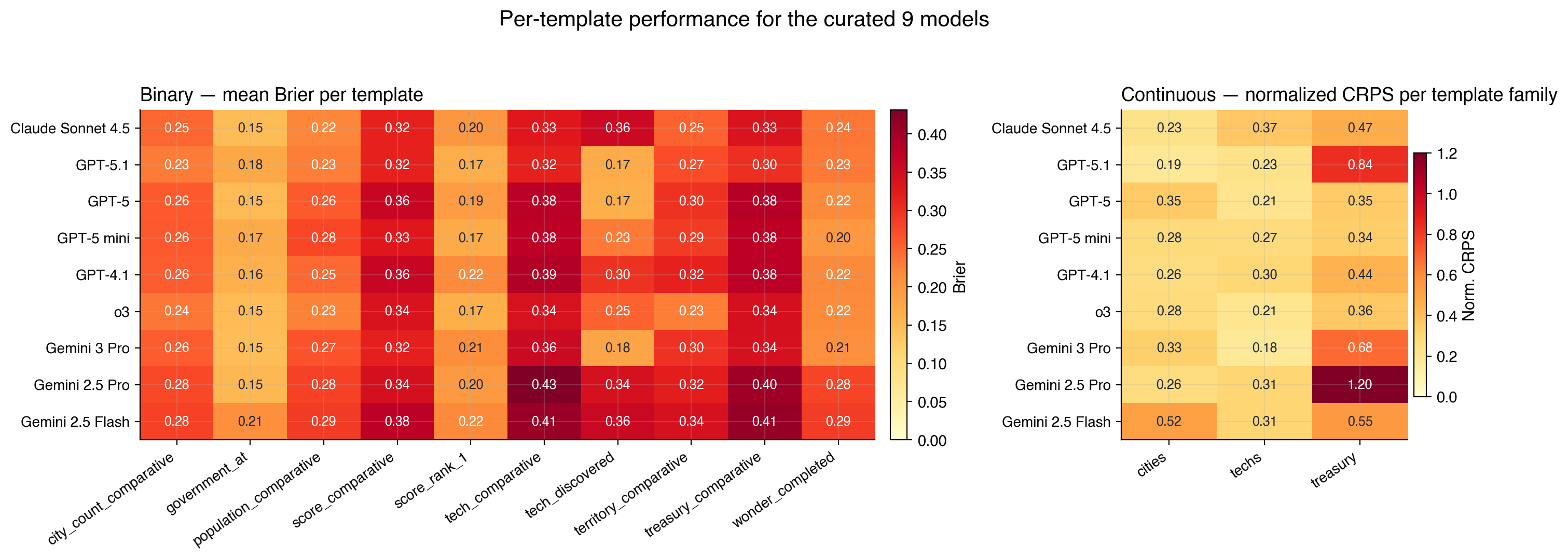}
  \caption{Template and horizon slices expose heterogeneous difficulty. This is useful for separating predictable quantities, such as technologies, from quantities more exposed to disruption, such as treasury or city count.}
  \label{fig:template-heatmap}
\end{figure}

\begin{figure}[H]
  \centering
  \includegraphics[width=.68\textwidth]{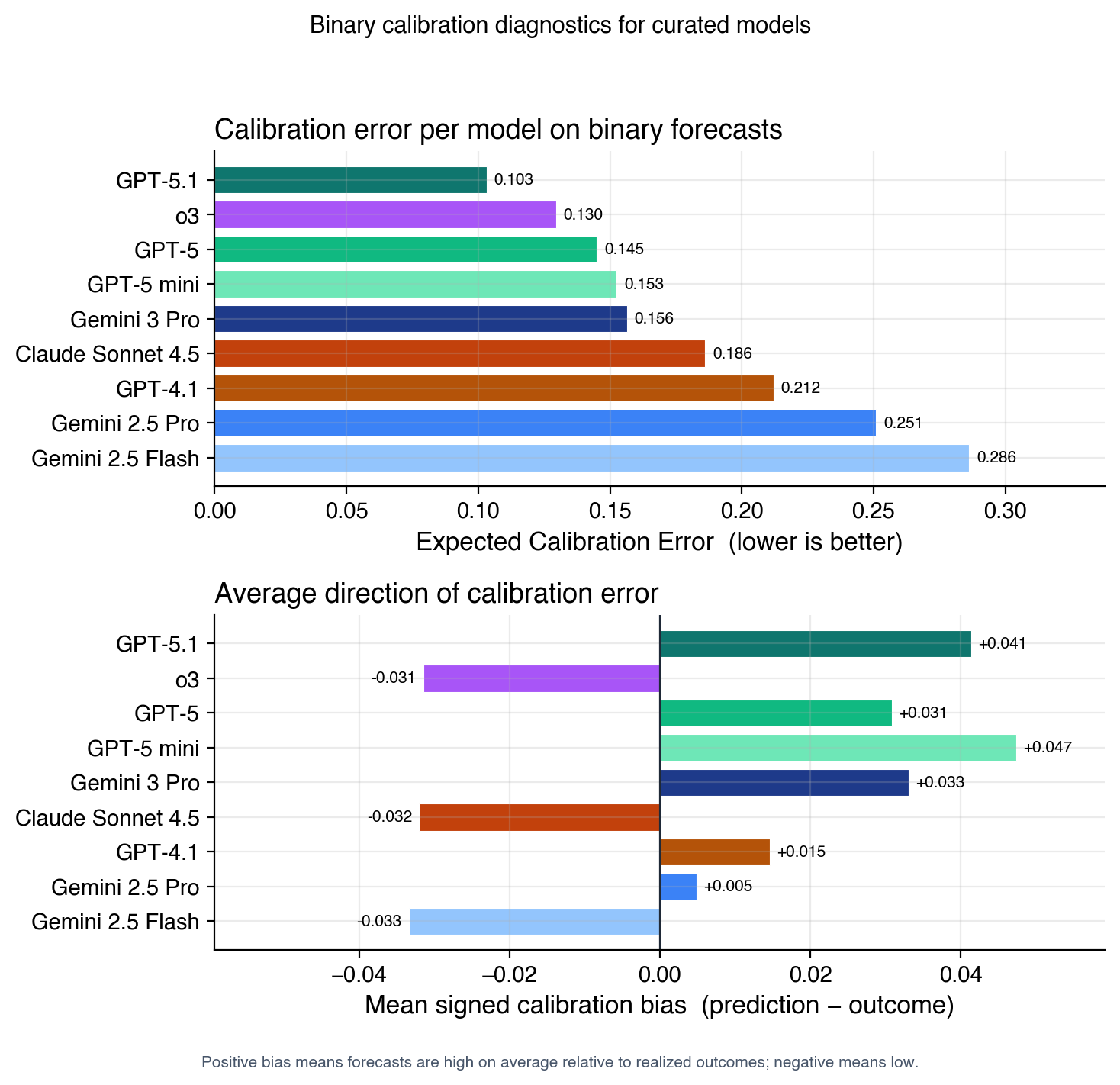}
  \caption{Binary calibration diagnostics for the curated model set. These diagnostics complement average Brier scores by showing whether errors are primarily calibration errors or discrimination errors.}
  \label{fig:binary-calibration}
\end{figure}

\clearpage
\section{Conditional and Intervention Diagnostics}
\label{app:conditional-interventions}

The archived conditional pilot runs use two interventions: a government switch to Republic and a +500 treasury shock. For each intervention, we compare a model's baseline forecast against its conditional forecast on the intervention-world outcome. Positive intervention gain means that conditioning on the intervention moved the forecast closer to the forked-world ground truth. These runs are not a final intervention benchmark, but they demonstrate the paired-rollout machinery that real-world forecasting benchmarks usually cannot provide.

\begin{figure}[H]
  \centering
  \includegraphics[width=.88\textwidth]{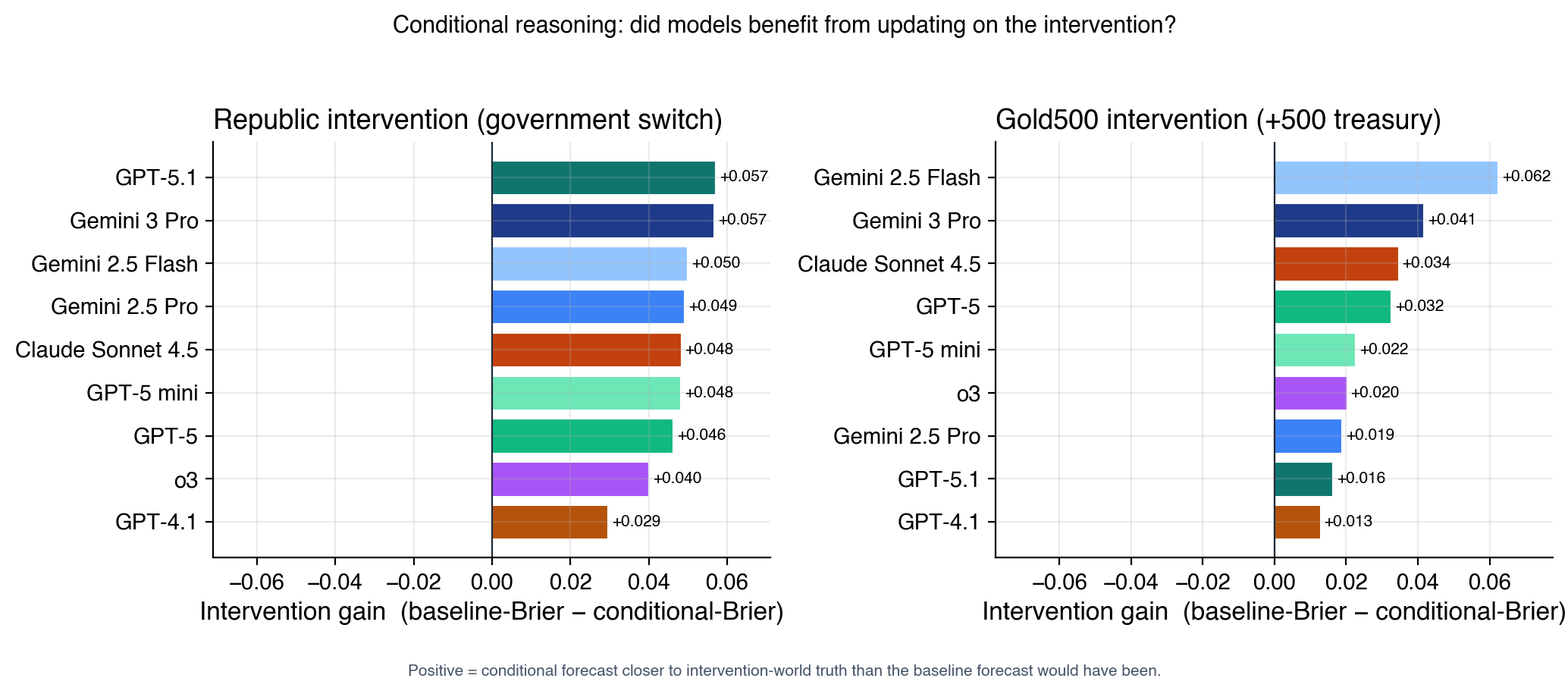}
  \caption{Conditional intervention gains for Republic and +500 gold pilots. Positive values indicate that the conditional forecast was closer to the intervention-world outcome than the baseline forecast would have been.}
  \label{fig:intervention-gap}
\end{figure}

The appendix also includes a small placebo control for conditional phrasing. In an archived Opus 4.5 Republic pilot, replacing the intervention with a null conditional leaves performance near baseline, while the real intervention sharply worsens it. This is the clearest current evidence that the score shift is tied to intervention content rather than only to the presence of an ``if'' clause.

\begin{figure}[H]
  \centering
  \includegraphics[width=.58\textwidth]{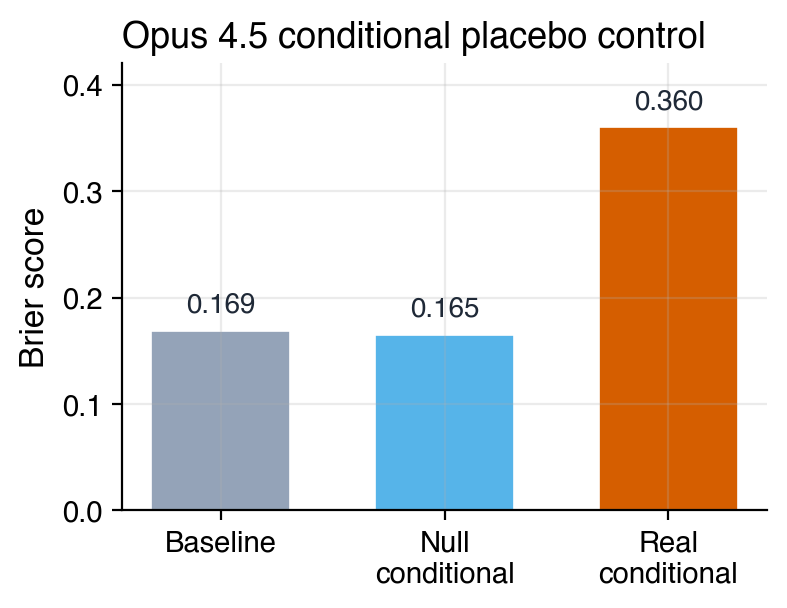}
  \caption{Placebo control for conditional framing in an Opus 4.5 Republic pilot. Replacing the intervention with a null conditional leaves Brier essentially unchanged from baseline, while the real intervention sharply degrades performance.}
  \label{fig:null-conditional}
\end{figure}

\clearpage
\section{Tail-Risk Calibration Example}
\label{app:tail-risk}

\begin{figure}[H]
  \centering
  \includegraphics[width=.82\textwidth]{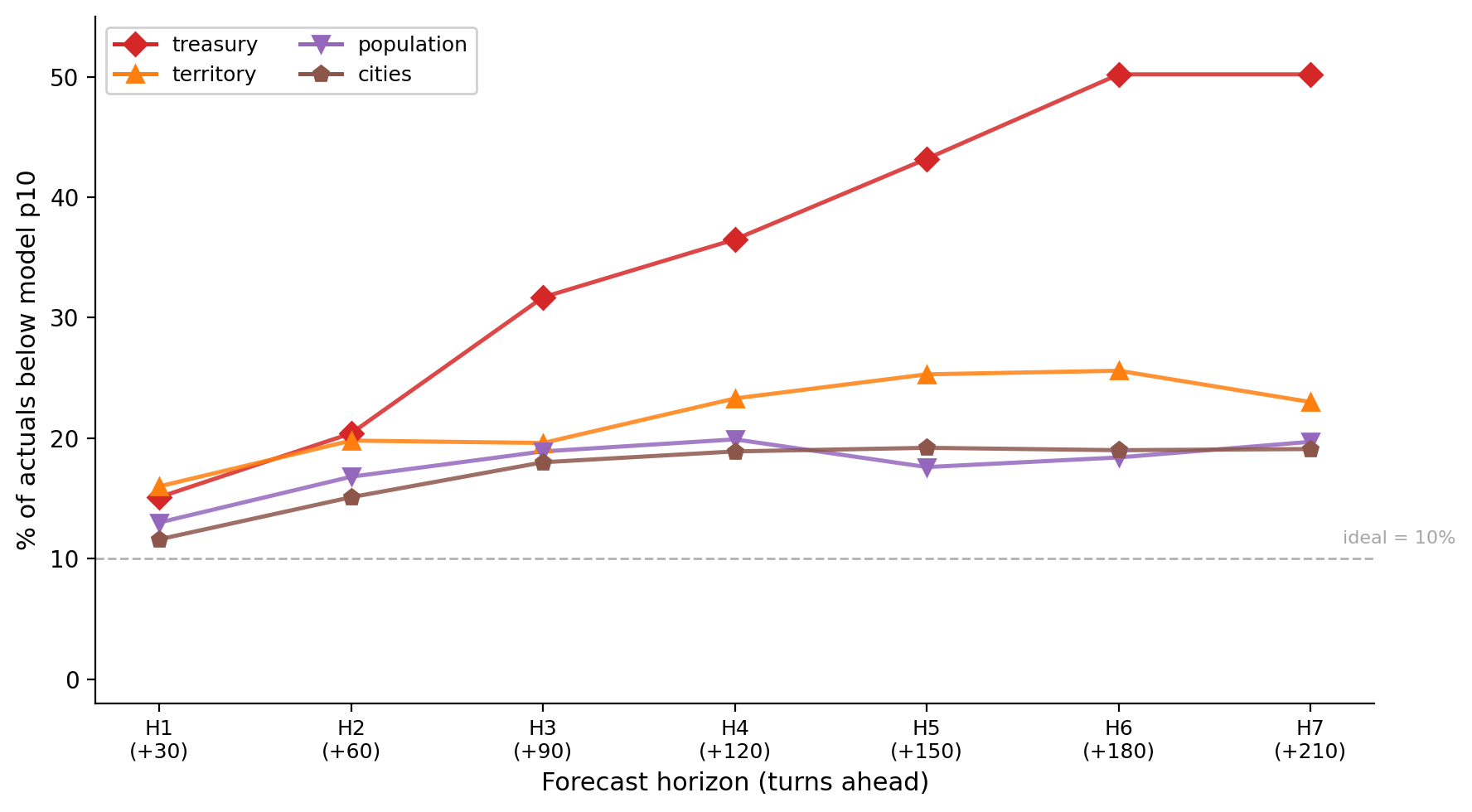}
  \caption{Example tail-risk diagnostic enabled by dense simulated-world sampling, adapted from a companion anonymous analysis on the same ForecastBench-Sim continuous question family \citep{anonymous2026capability}. On disruptable continuous templates, the fraction of actuals falling below the model's stated p10 rises well above the nominal 10\% rate, with treasury reaching about 50\% by H6--H7.}
  \label{fig:tail-risk}
\end{figure}

The same immediate-resolution substrate can support downstream studies of rare and disruptive outcomes. \Cref{fig:tail-risk} shows one example adapted from a companion anonymous analysis on the same ForecastBench-Sim continuous question family \citep{anonymous2026capability}: once many simulated worlds are sampled, the benchmark can expose systematic undercoverage of downside outcomes by horizon and template.

\end{document}